\documentclass[10pt,twocolumn,letterpaper]{article}

\usepackage{cvpr}
\usepackage{times}
\usepackage{epsfig}
\usepackage{graphicx}
\usepackage{amsmath}
\usepackage{amssymb}
\usepackage{soul}
\usepackage{capt-of}
\usepackage{multirow}
\usepackage{adjustbox}
\usepackage[thinlines]{easytable}
\usepackage{gensymb}

\usepackage[breaklinks=true,bookmarks=false]{hyperref}

\cvprfinalcopy 

\def\cvprPaperID{8308} 

\DeclareMathOperator*{\argmax}{arg\,max}

\DeclareMathOperator*{\subjectto}{subject\,to}

\begin{document}


\author{Neil Fendley, Max Lennon, I-Jeng Wang, Philippe Burlina, Nathan Drenkow\\
The Johns Hopkins University Applied Physics Laboratory\\
11100 Johns Hopkins Rd\\
Laurel, MD USA\\
{\tt\small {nathan.drenkow}@jhuapl.edu}
}

\title{Jacks of All Trades, Masters Of None: \\ 
Addressing Distributional Shift and Obtrusiveness \\
via Transparent Patch Attacks}
\maketitle

\begin{abstract}
\vspace{-.4cm}
We focus on the development of effective adversarial patch attacks and -- for the first time -- jointly address the antagonistic objectives of attack success and obtrusiveness via the design of novel semi-transparent patches.  This work is motivated by our pursuit of a systematic performance analysis of patch attack robustness with regard to geometric transformations.  Specifically, we first elucidate a) key factors underpinning patch attack success and b) the impact of distributional shift between training and testing/deployment when cast under the Expectation over Transformation (EoT) formalism. 
By focusing our analysis on three principal classes of transformations (rotation, scale, and location), our findings provide quantifiable insights into the design of effective patch attacks and demonstrate that scale, among all factors, significantly impacts patch attack success. 
Working from these findings, we then focus on addressing how to overcome the principal limitations of scale for the deployment of attacks in real physical settings: namely the obtrusiveness of large patches. Our strategy is to turn to the novel design of irregularly-shaped, semi-transparent  partial patches which we construct via a new optimization process that jointly addresses the antagonistic goals of mitigating obtrusiveness and maximizing effectiveness. 
Our study -- we hope -- will help encourage more focus in the community on the issues of obtrusiveness, scale, and success in patch attacks.

\end{abstract}

\begin{figure}[t!]
\begin{center}
    \begin{tabular}{c}
        \includegraphics[width=0.6\columnwidth]{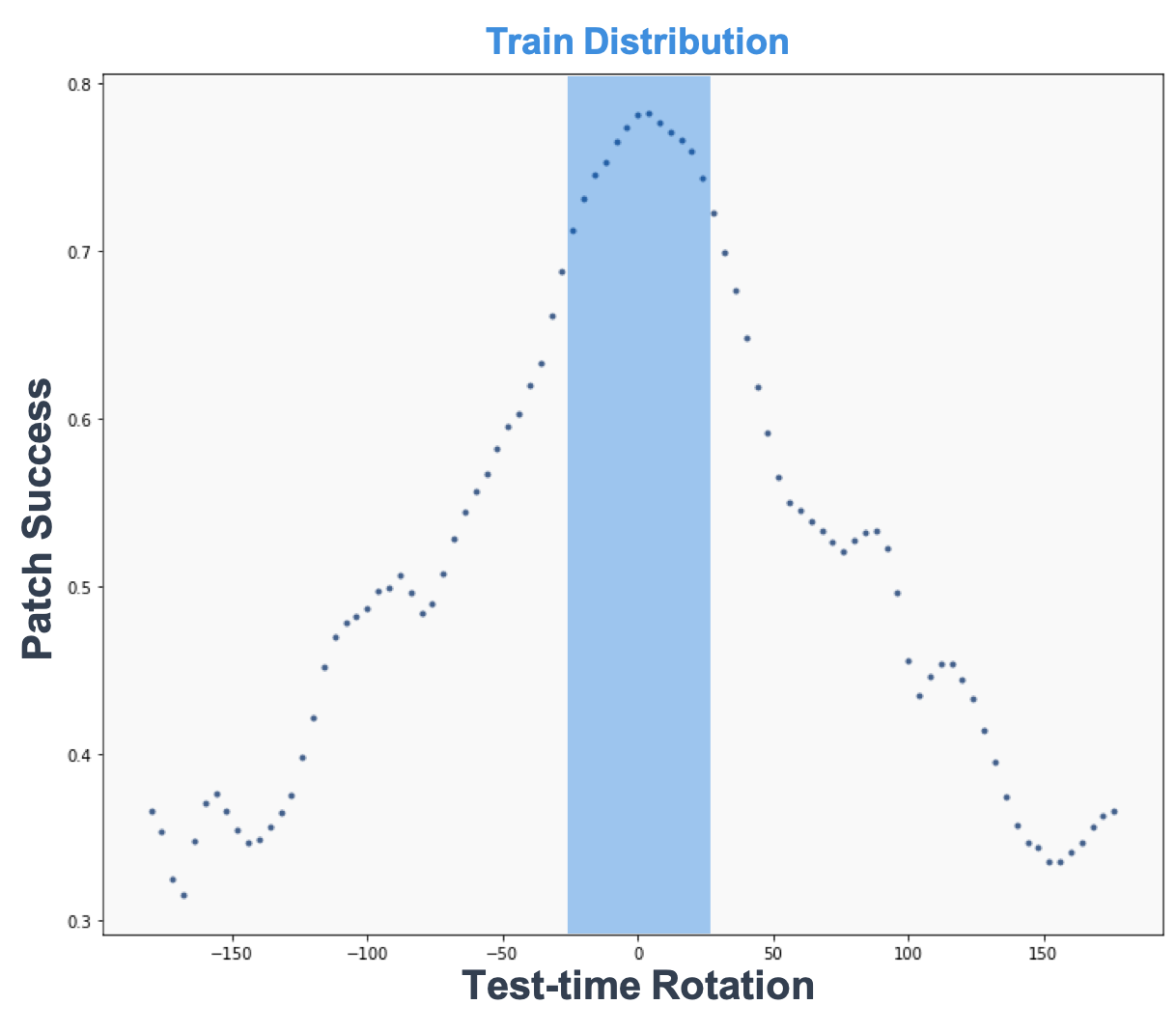} \\
        \includegraphics[width=0.7\columnwidth]{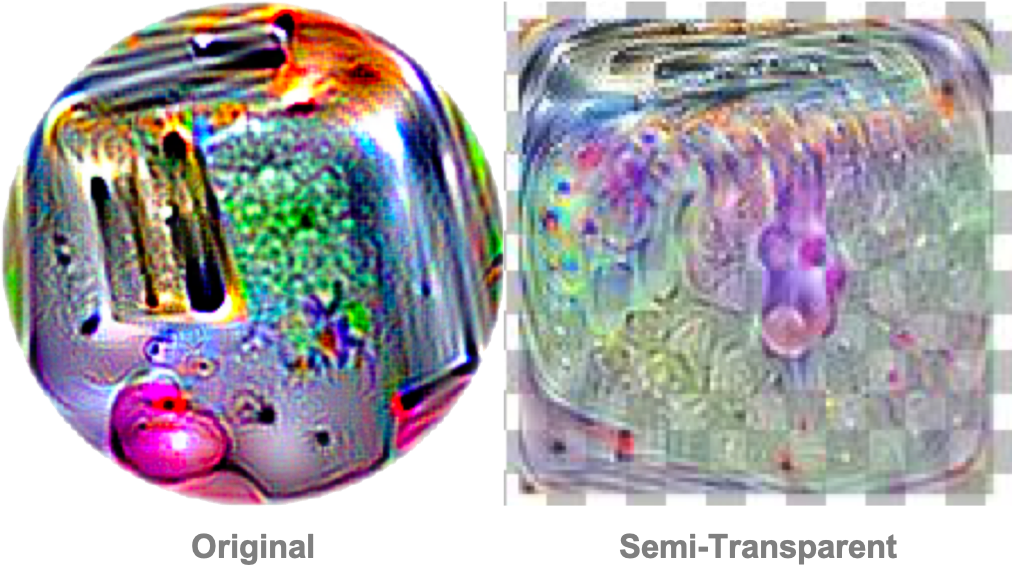}
    \end{tabular}
\end{center}
   \caption{[\textbf{top}] Demonstrating the impact on patch attack success for test-time transformations sampled out-of-distribution. 
   [\textbf{bottom}] A novel method for generating semi-transparent patch attacks to address the balance between patch scale, obtrusiveness, and success. (Toaster patch used above from~\cite{brown2017adversarial})
   }
\label{fig:summary}
\end{figure}

\section{Introduction}
Deep learning (DL) has demonstrated performance seemingly on par with humans for various problems such as medical image diagnostics~\cite{esteva2017dermatologist},  game playing~\cite{mnih2015human}, and other tasks~\cite{lecun2015deep}. However this apparent success has been met with the prospect~\cite{goodfellow2014explaining} that deep networks are very fragile to adversarial attacks, which motivated many studies in the past several years researching methods for attacks and defenses, and the corresponding significant growth of interest in the community in the field of {\it adversarial machine learning} (AML) specifically applied to DL models. Several approaches to AML as applied to DL have been envisioned, and a simple taxonomy useful to frame the goals for this work  consists of distinguishing between a) generating attacks that affect an entire image, are designed to work on a specific image, and are carried out by additive non-perceivable perturbations (such as in~\cite{szegedy2013intriguing},~\cite{goodfellow2014explaining},~\cite{moosavi2016deepfool},~\cite{moosavi2017universal}), vs. b) attacks that are confined to a specific sub-window of the image and are designed to affect a wide set of images and classes of objects (i.e., a patch attack). 

Attacks that are of interest for this work use the approach in (b). These attacks are more suitable to be implemented in the physical domain since they can be printed on contiguous surfaces and placed more easily in a scene -- a significant concern for applications such as automotive and robotic autonomy and related areas. The first successful design of such an attack was reported in~\cite{brown2017adversarial} where it was demonstrated that an adversarial patch can be designed by using a loss function that includes a term that expresses an expectation over geometric transformations including rotation, translation and scale. This was based on work originally reported in~\cite{athalye2017synthesizing}. 


Additionally, and to better frame our motivations here, when it comes to patch attacks for physical scenes, one must distinguish between the specific case (where attacks are generated and optimized for a particular scene) and the general case (where attacks are generated without any prior knowledge of the scene/context in which they will be placed).  In the former scenario, patch patterns and placement may be optimized more highly given greater knowledge about the attack scene.  However, we focus on the latter scenario which makes fewer assumptions about the scene and is thus likely to produce attacks that generalize more effectively.  We wish to understand the robustness of deep neural networks to attacks developed for these scenarios and then design the most effective attacks under these assumptions.


\begin{figure}[t!]
\begin{center}
    \begin{tabular}{c}
        \includegraphics[width=0.75\linewidth]{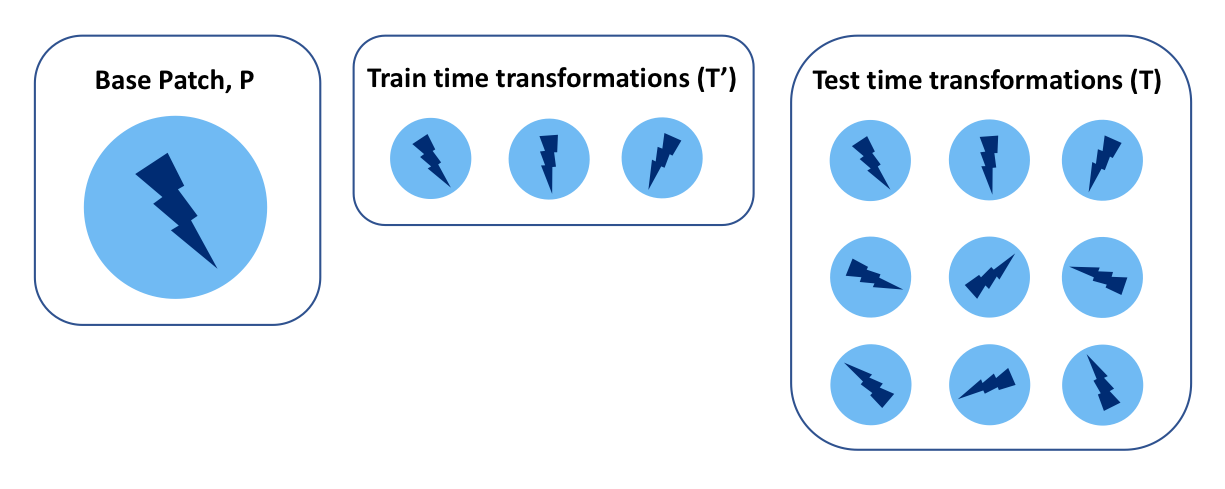}
    \end{tabular}
\end{center}
   \caption{Constrained EoT limits the range of transformations (e.g., rotation angles, scales, etc.) for optimizing the patch, but increases the range used at test time.  Test time transformations are assumed to be unconstrained as this is consistent with many real world use cases.}
\label{fig:constrained_eot}
\end{figure}

\section{Contributions}
The novel contributions of this are as follows: a) we present a principled methodology for the evaluation of patch attacks and the train/test-time factors that impact their success.  In particular, we study, under the framework of the expectation over transformation approach, the impact of distributional differences between patch optimization and deployment conditions and their subsequent effect on patch attack success.  This study allows new insights into factors leading to attack success and in particular demonstrate that among all -- patch scale is a driving factor for success, and that rotation factors suffers from a ``jack of all-trades, master of none'' pathology. b) Armed with these observations we next consider the question of how to best design effective patches. That is, patches that scale up but still retain desirable factors with regard to deployment and detectability (i.e. unobtrusiveness). To our knowledge we report the first ever design of semi-transparent patches that address these objectives. We also develop a novel c) an optimization framework that results in the design of such patches and d) develop new methods to characterize effectiveness in this new scale/obtrusiveness/success trade space, and, given scale as a key limiting factor, we develop a novel measure for patch obtrusiveness. 


\section{Analysis of Expectation over Transformation}
\subsection{Motivation}
This paper focuses on patch-based attacks initially inspired by the EoT method first introduced by~\cite{athalye2017synthesizing}.  Early experiments with EoT produced results as shown in Figure~\ref{fig:summary}.  In particular, it could be observed that when the test-time transformation distribution is unconstrained and mismatched with the train-time distribution, the patch attack severely under-performs.  Furthermore, as we seek to optimize new patches, there has been limited examination of how to define the patch transformation distributions used in the optimization.  

To dig deeper into these issues, we performed a systematic study of the effects of standard patch transformations on attack success.  Specifically, we identify how train and test-time distributions (and mismatches between them) relate to patch effectiveness.  Given that EoT is the standard method for generating patch attacks, our intent is not to disprove the method's utility but rather to provide quantifiable evidence and practical insights regarding its usage.

\subsection{Expectation over Transformation}\label{sec:eot}
\label{sub:EoT}
In one of the most common formulations of the problem, an adversarial example is generated by perturbing a clean input image $x$ (taken from the set of all possible images $X$) to produce an adversarial exemplar image, $x'$, such that a classifier will produce a correct classification, $y$ (taken from a set of $Y$ classes) for $x$, but an incorrect classification, $y'$, for $x'$.  Following~\cite{athalye2017synthesizing}, the perturbation is determined more formally via optimization:
\begin{equation}
\label{eq:ae}
    \begin{array}{l}
    \argmax_{x'} P(y'|x') \\
    \subjectto \|x'-x\|_p < \varepsilon 
    \end{array}
\end{equation}
In the case of a targeted attack, the optimization is designed to find the $x'$ that maximizes the likelihood of producing the targeted label, $y_t'$.

As stated in~\cite{athalye2017synthesizing}, this approach fails to account for real world transformations and was thus extended by including such transformations directly in the optimization process.  The result is a modified optimization:
\begin{equation}
\label{eq:EoT}
    \begin{array}{l}
    \argmax_{x'} \mathbb{E}_{t\sim T} P(y'|t(x')) \\
    \subjectto\mathbb{E}_{t \sim T}[d(t(x), t(x'))] < \varepsilon 
    \end{array}
\end{equation}
where $T$ is a distribution of transforms, $d()$ is a distance measure, and $t(x)$ is a sampled transform applied to the image $x$.  

This approach was further modified by~\cite{brown2017adversarial} to produce patch-based attacks (where perturbations are spatially co-located) robust to transformations likely to occur in the physical world. In particular, that work used a patch application operator, $A(p, x)$, which applies transformations and perturbations within a confined spatial region.  The result is the following:
\begin{equation}
\label{eq:patch_EoT}
    \argmax_{p} \mathbb{E}_{x\sim{X}, t\sim{T}} P(y'|A(t(p), x))
\end{equation}
Without loss of generality, the focus is on transformations $T$ including distributions over location, scale, and rotation.

While the above methods demonstrate the advantage of expanding the optimization to incorporate potential real-world transformations, the trade-off between attack success and transformations considered has not yet been investigated and is the focus of our study.

\subsection{Constrained Expectation over Transformation}
The main endeavor of this section is to determine how constraints imposed on the distribution of transformations during the patch optimization affect success when transformations are unconstrained at test time.  For physical attacks, this is a key question since the attacker can optimize and place the patch in the scene (under certain transformation and distributional assumptions) but has no control over the pose of the system-under-attack, which may present a distributional shift at attack time. In~\cite{athalye2017synthesizing} and~\cite{brown2017adversarial}, the transformations were limited to a small range of rotation angles and scales. Here we ask whether there is any trade-off or negative impact on the effectiveness of EoT if the range of transformations is expanded or mismatched between optimization and testing.

We run a series of experiments (described in Section~\ref{sec:EoT_experiments}) with modifications to (\ref{eq:EoT}) whereby during the optimization we consider the distribution $T$ (typically, uniform) over a compact support $\mathcal{T}$ (Figure~\ref{fig:constrained_eot}).  At test time, we then use the full support of $T$ to assess patch success.  We first perform these experiments considering scale and rotation transformations separately, then repeat with scale and rotation jointly optimized. 

The motivation for this formulation and set of experiments is three-fold.  First, by creating a difference between train and test-time transformation distributions, we can assess whether the added constraint improves or degrades patch success. For instance, we can quantify trade-offs or benefits that might exist for optimizing over the full range of rotation angles relative to a smaller number.
Additionally, we can further study in isolation the effects of individual transformations on patch effectiveness. Lastly, we can leverage insights gained to more appropriately constrain patch optimization relative to specific applications.
We note that while we limit our experiments to examining patch rotation, scale, and location, this evaluation can be generalized to other relevant transformations under consideration.

\section{Experiments}
\label{sec:EoT_experiments}
\subsection{Base experiment}
\label{sub:base_experiment}
All of the following experiments are derived from a common base experiment.  Since the primary focus of this section is on providing a quantitative analysis of the effects of testing transformations outside the distribution of training images, we restrict our experiments to a white-box scenario where the model and architecture used for patch optimization is the same as the one that undergoes the patch attack. We view this as a best case scenario in light of the challenges with attack transferability (as discussed in~\cite{demontis2019adversarial}~\cite{liu2016delving}~\cite{papernot2016transferability}~\cite{zhou2018transferable}). All experiments in this paper optimize and test patch attacks using a ResNet50~\cite{he2016deep} model trained on ImageNet.

In each of the next experiments, we use the following procedure.  First, we select the transformation of interest and define its corresponding train time support $\mathcal{T}$ (e.g., rotation angle $\theta\in[0,45]$). Then, we sample 100 classes uniformly from the standard ImageNet classes to be our target labels and for each class, we optimize a patch attack following~\cite{brown2017adversarial}.  This attack is then tested by sampling a test-time configuration from the full support of the transformation distribution, $T$, for each image in a set of 512 randomly sampled from the ImageNet test set.  Patch locations during optimization and at attack time are randomized (unless specified otherwise - Section~\ref{sub:location_experiments} examines this transformation separately).

\subsection{Scale}
\label{sub:scale_experiments}
{\bf Question:} The scale of patch attacks in the real world may be the most challenging factor to control at attack time.  Thus, we ask: {\it what is the impact of scale on patch success? Will patches optimized at small scales maintain effectiveness at large scales (and vice versa)?}

{\bf Approach:} These questions address whether the patch optimization process is producing patterns that rely on fine-grained details or embody general yet robust attack patterns.  To address the scale questions, we optimize the set of patches following the base experiment setup with patch rotation fixed at $0\degree$ and location randomized.  We perform two variants of the experiment. The first experiment starts at a nominal scale, $s_o\in[s_{min},s_{max}]$, and bounds the support $\mathcal{T}$ such that scale varies from $s_o$ up to a maximum scale $s_{max}$ (i.e., $[s_o, s_{max}]$).  The second experiment starts from a nominal scale and optimizes from the minimum up to nominal (i.e., $[s_{min}, s_o]$).  Both experiments allow the full support at test time with scale sampled from the entire distribution $T\sim\textit{U}(s_{min}, s_{max})$.  

\begin{figure}[t!]
\begin{center}
    \begin{tabular}{c}
        \includegraphics[width=0.98\columnwidth]{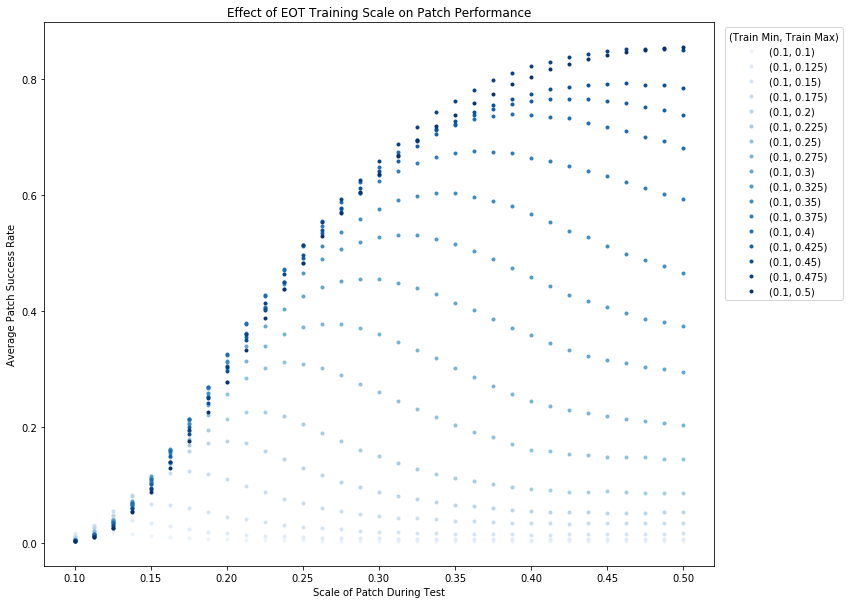} \\
        \includegraphics[width=0.98\columnwidth]{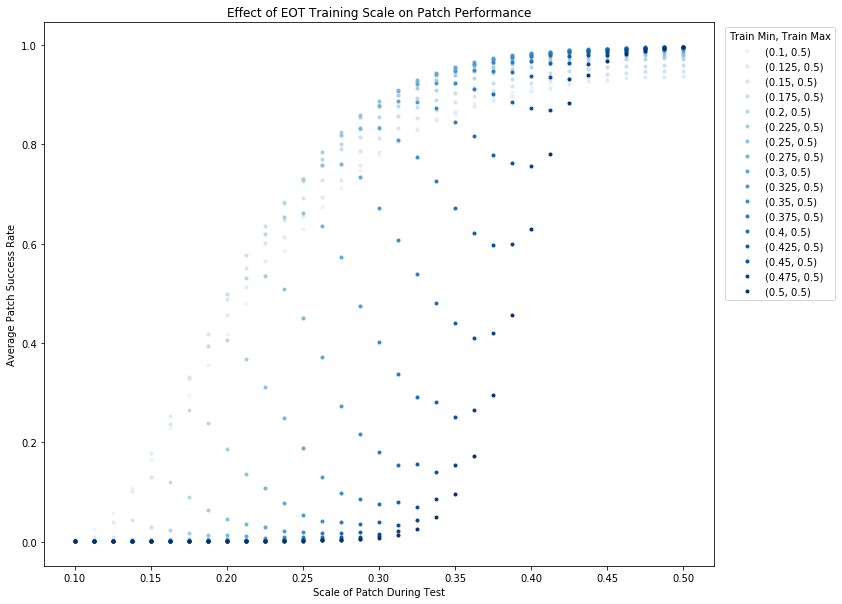}
    \end{tabular}
\end{center}
\caption{Results for EoT with variable scale.  (top) Shows how patch performance changes when the minimum scale is fixed and the maximum scale varies.  (bottom) Shows performance differences starting from a fixed maximum size and allowing the minimum scale to vary.  Shades correspond to the range of scales represented in the expectation.  Both plots show attack success rate, so higher is better.}
\label{fig:scale_exp}
\end{figure}

{\bf Experiments:} Results for both experiments are captured in Figure~\ref{fig:scale_exp}.

{\bf Discussion:} Both experiments indicate that patch success is highly dependent on scale.  We find that patches optimized at larger scales are more likely to have success at small scales than patches optimized in the reverse direction.  We attribute this to the fact that a higher resolution for the patch forces the optimization to place less importance on individual pixels and to reveal patterns which are preserved when forced to lower resolutions. Regardless of the optimization approach, the results clearly indicate significant differences in success rate between large and small patches, which suggests that additional techniques may be necessary to overcome this practical limitation (a topic addressed in Section~\ref{sec:partial}).  

\subsection{Rotation}
\label{sub:rotation_experiments}
{\bf Question:} We next seek to understand how robust patch attacks are to out-of-distribution rotations experienced at test time.  We ask the following question: {\it are patch attacks more robust when the patch is optimized over a larger range of rotations? how does patch success modulate with respect to this range?}

{\bf Approach:} We again run the base patch optimization procedure now holding scale fixed while allowing location to vary randomly. However, during patch optimization, the rotation distribution is fixed such that angles are selected from a bounded support, $\mathcal{T}$, with angles taken from $[-\theta,\theta]$.  We run a series of variants of the base experiment by varying $\theta$ at intervals of $\pi/5$.

\begin{figure}[t!]
    \centering
    \begin{tabular}{c}
       \includegraphics[width=0.98\columnwidth]{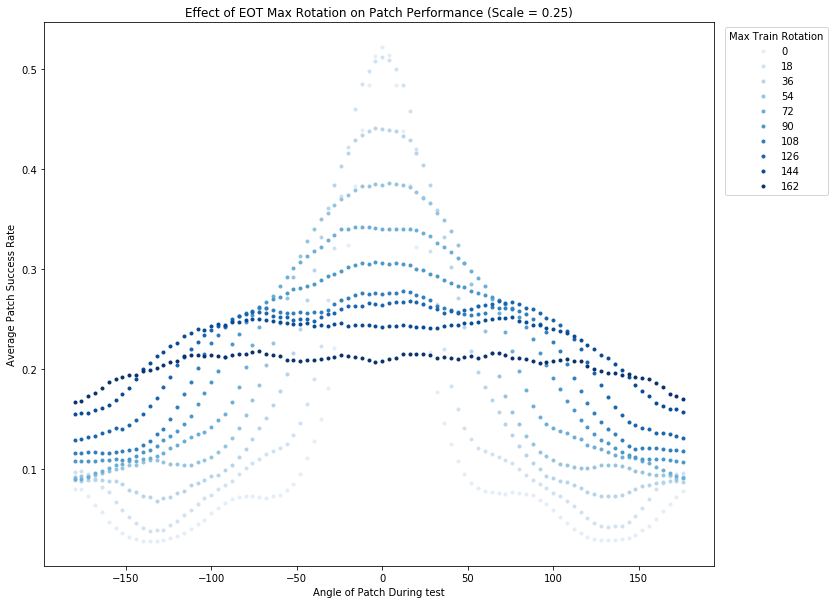}
    \end{tabular}
    \captionof{figure}{Results for EoT when only angles are considered. The plot shows patch success (higher is better) at test angles from $[-\pi,\pi]$. Each shade of blue represents the maximum angle allowed during patch optimization (darker shades imply more angles considered).}
    \label{fig:rotation_exp}
\end{figure}

{\bf Experiments:} Results of the rotation experiments are captured in Figure~\ref{fig:rotation_exp}.

{\bf Discussion:} The rotation experiments illustrate a jack-of-all-trades, master-of-none result whereby patches optimized over a wider angle distribution exhibit lower peak performance yet higher average performance.  One possible explanation for this result is that the conventional formulation of EoT focuses on finding a patch that maximizes the target class probability in expectation rather than including the rotation transformation directly in the optimization.  While the loss is computed over all rotations sampled during a given training batch, the patch is always updated in its canonical orientation.  When taking the expectation, it is likely that local patch regions compete with each other at specific angles.  As the range of candidate rotations increases, the EoT forces the local regions to become less competitive so that they are more effective over a range of angles.  However, the undesired effect is that the local region patterns may have lower performance at specific angles.


\subsection{Location}
\label{sub:location_experiments}
{\bf Question:} We perform experiments to explore the incidence of the location of adversarial patches in the image on the attack performance. Specifically, {\it we probe the veracity of the hypothesis that placement of masks at locations of maximum impact  results in maximum attack effectiveness (where, as  we shall explain next,  maximum impact is taken to be maximum saliency).}

{\bf Approach:} We must first clarify again that we're considering the general case of patch attacks whereby we optimize the patch making assumptions about the types and range of patch transformations, but make no assumptions about the scene or context in which the patch attack is applied.  We make this distinction because in the specific attack scenario, a patch and location may be optimized jointly (or conditionally) if certain characteristics of the scene are assumed.  In this experiment, we focus on the impact of the presence/absence of systematic biases in the location selection which may impact patch success.  We utilize saliency (as determined via white-box methods) to identify how a patch's placement in regions of high/low importance to the deep network affect the patch success.

We compute saliency via conventional computation methods that use gradient descent to find locations in the input image domain that result in maximum softmax output for the image class label. The basic hypothesis driving this choice for location is that, by placing the patch over maximally salient locations in the image, the patch will gain maximum attack potency by both a) occluding the most important cues leading to the correct label and b) replacing them with cues targeting the new targeted class.

\begin{figure*}[t]
    \centering
    \begin{tabular}{c}
      \includegraphics[width=0.8\textwidth]{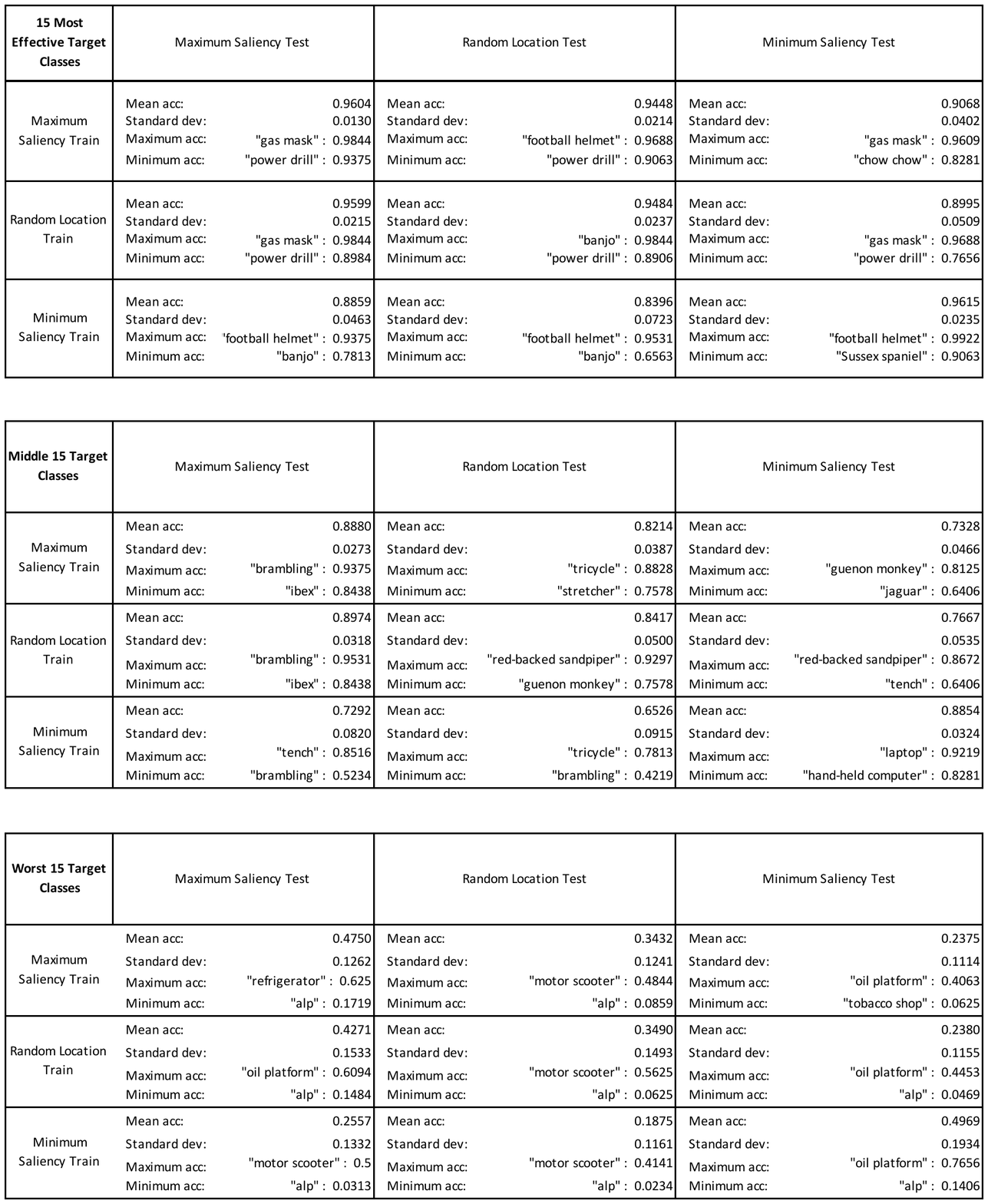}
    \end{tabular}
    \vspace{-5cm}
    \captionof{figure}{Results for location: showing the 15 most effective target classes (top), median (middle) and least (bottom).}
    \label{fig:location_exp}
\end{figure*}

{\bf Experiments:} For EOT-based training and testing, we consider three location strategies, corresponding to minimum and maximum loci in the saliency map as well as random location. We then explore the results of attacks using masks by combining the scenarios of \{min, max, random\} train location with \{min, max, random\} test locations. The results are shown in tables given in Figure~\ref{fig:location_exp}. We observe a strong dependence between patch success and target label. While we view this as an important relationship to study, for now we simply capture it by comparing patch performance binned according to the top, middle and bottom performing target classes. For each situation we report the mean accuracy when computed on a set of images along with additional statistics (standard deviation, as well as min and max accuracy).






{\bf Discussion:} Looking at the results we found that the highest effectiveness is obtained in general when selecting the patch placement at test time according to the same rules as during the optimization (e.g., random location in both). There is a slight gain in the patch success when placing it in regions of high saliency, but it is not statistically significant, and  the preference on placing the patch in the same type of location as it was trained seems to trump other considerations.  

On the other hand, if we train randomly, and place the patch anywhere at attack time, we don't observe much of a performance hit.
The fact that maximum placement does not over-perform reveals results that are somewhat counter to what one would hypothesize and suggests an 'easy-does-it' strategy, namely that you don't have to calculate maximum saliency at training time and place patches at optimal locations at attack time. Instead, you can work with random locations at train and attack time and still achieve results that are nearly as good as the maximal location.


In sum, location placement is one area in which -- unlike rotation -- adopting a lazy approach pays off.  While this result was confirmed in this last set of EoT experiments, it helps to justify the choice of random location used for the scale and rotation variants of the base experiment.

\section{Improving Patch Attack Effectiveness and Unobtrusiveness}
\label{sec:partial}

We observed in Section~\ref{sub:scale_experiments} the dominant effect of scale on patch effectiveness, specifically noting a strong drop-off in performance as the patch image dimension decreases below 30\% of the image edge size.  This leads us to ask the question: {\it can we devise a strategy to increase the patch size (and subsequent effectiveness) without simultaneously increasing the observability of the patch itself?}

\subsection{Approach} 
First, we must clarify two things. As before, we make the assumption that the attacker has no control over the camera pose and so cameras can always be positioned to force the patch to be arbitrarily small.  However, we aim in this section to devise a method to increase the patch size to improve overall patch success.  

Second, while we frame the question as one of observability, we're not asking it from the standpoint of measuring limits of human perception.  Instead, we view observability as roughly equated to scene occlusion.  To address the motivating question, we devise a measure of \textit{patch obtrusiveness}. Since a primary advantage of patch-based attacks is the ability to print contiguous yet semi-transparent patterns, we can define patch \textit{obtrusiveness} as a measure of the overall opacity of the patch.  Fully opaque patches have maximal obtrusiveness as they are most easily identified at test time relative to a nearly transparent patch. Furthermore, our patch {\it obtrusiveness} measure creates a trade space whereby scale, opacity, and success rate can be optimized together according to the desired operating point.

More precisely, we measure opacity by first defining a mask, $M$, which has the same shape (namely $h \times w$ for rectangular patches) as the patch $P$ and where $M_{ij}\in[0,1]$ for each $i,j$ in $M$.  From here, we define \textit{patch obtrusiveness (PO)} as:

\begin{equation}\label{eqn:obt}
    PO(M) = \frac{\textit{total mask value}}{\textit{total mask area}} = \frac{\sum_{i,j}M_{ij}}{hw}
\end{equation}

This metric allows us to create some measure of equivalence between small, opaque patches and large, partially transparent patches.  When $PO \approx 0$, the patch is essentially transparent and maximally subtle in the image.  When $PO=1$, the patch is fully opaque and is equivalent to the current conventional output from the EoT method.  Ultimately, our goal is to side-step the scale limitations observed with EoT (as discussed in Section~\ref{sub:scale_experiments}) and produce larger, more effective patches without increasing observability.

Given our definition of $PO(M)$, we now include the mask $M$ as part of the optimization.  Whereas standard EoT applies a sampled transform $t(p)$ to patch $p$ and then applies the transformed patch directly over the image $x$ (i.e., $A(t(p), x)$), we modify the application function $A()$ to blend the patch and the image according to the values of $M$.  Namely, the adversarially attacked image is created as follows: For all $i,j$ in $x$, 
\begin{equation}
    \begin{split}
        x'_{ij} & = A\left(t(p), x; t(M)\right)_{ij} \\
            & = t(M)_{ij} * t(p)_{ij} + \left(1 - t(M)_{ij}\right) * x_{ij},
    \end{split}
\end{equation}
\noindent
where $x$ is the clean image, $M$ is the optimized mask, $p$ is the optimized patch, and $t$ is the sampled transformation to apply equally to the patch and mask.  We assume that $t(p)$ produces an image of the same dimension as $x$ where $p$ has been transformed and placed at a location also valid in $x$.

Now that we can apply $M$ to the patch to attack a clean image, we optimize the patch $p$ and mask $M$ simultaneously:

\begin{equation}
\label{eq:partial_patch_EoT}
    \argmax_{p,PO(M)}~\mathbb{E}_{x\sim{X}, t\sim{T}} P(y'|A(t(p), x; t(M)))
\end{equation}

For targeted attacks, the optimization loss term for a single attacked image is:

\begin{equation}
    \begin{split}
        L_{total} &= L_{target} + \gamma~L_{PO} \\
        L_{target} &= CE(y',y_{target}) \\
        L_{PO} &= (PO(M))^2 
    \end{split}
\end{equation}
where $y'$ is the label produced by the network for the attacked image $x'$, $CE$ is the standard cross-entropy loss, and $PO(M)$ is the patch obtrusiveness defined in (\ref{eqn:obt}).  Gradients with respect to the loss term are backpropagated to both the patch and mask pixels so as optimize them jointly. Due to differences in the natural scale of the loss terms, we include $\gamma$ to allow more control over the competition between patch obtrusiveness and its ability to fool the attacked network. 

\subsection{Experiment} 
\label{sub:obtrusive_exp}
Following the base experiment approach described in Section~\ref{sub:base_experiment}, a series of patches (and associated masks) are optimized and tested over a sampling of target labels and test images. We hold rotation fixed and use random locations during patch optimization and evaluation.  The rotation assumption can be viewed as a best case given the results in Section~\ref{sub:rotation_experiments} and the location strategy is based on the results in Section~\ref{sub:location_experiments}.

Given $\gamma$ is designed to control the trade-off between patch success and obtrusiveness, we employ a curriculum learning-like procedure whereby minimizing patch obtrusiveness is prioritized initially then down-weighted later on to encourage greater patch performance.  Inspired by~\cite{wang2019neural}, we change the $\gamma$ value throughout the patch optimization process depending on the value of the loss.  The $\gamma$ term is held fixed until the loss remains below a threshold (e.g., nominally 0.1) for a fixed patience period (e.g., min 5 iterations). After the patience period, the $\gamma$ value is decreased (according to its schedule) and the process repeats.  We observe during training that changes in $\gamma$ usually produce a spike in the loss, so the patience period promotes a certain degree of stability and is critical to achieving high-performing semi-transparent patches.


\begin{figure}
    \centering
    \begin{tabular}{c}
       \includegraphics[width=0.9\columnwidth]{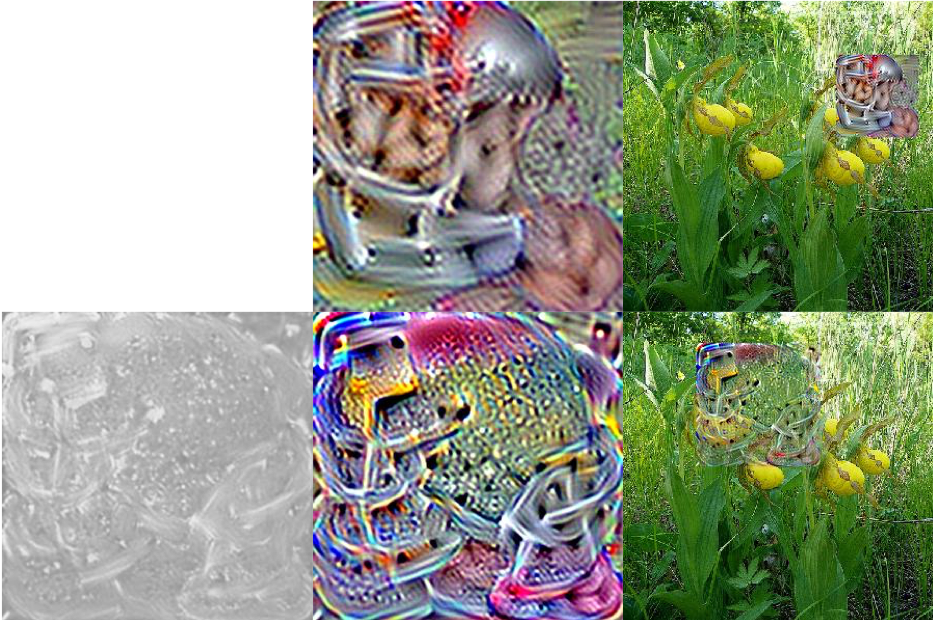}
    \end{tabular}
    \captionof{figure}{Example of patch attacks with and without optimization to balance obtrusiveness and success.  Top row is the control patch and its application to a test image.  The bottom row shows the mask, the full patch, and the application of both to the same test image.}
\label{fig:mask_example}
\end{figure}

Figure~\ref{fig:mask_example} illustrates a standard EoT-based patch attack as well as an example produced with our approach showing a sample mask, full patch, and application of both to attack a clean image.

To test our method, we follow the base experiment setup but expanding to 1200 iterations during optimization (i.e., until convergence) and employing a learning rate of 5.0.  We fix the scale between [0.4, 0.5] and then run the optimization.  At the end of the process, the final obtrusiveness value is recorded and the patch is tested on a random set of 512 held-out images.

For the control experiments, we take the final obtrusiveness score and label for the semi-transparent patch and produce a fully-opaque control patch at a scale that achieves the same obtrusiveness score. Note that because the control patches have no transparency, they are naturally smaller.  The control patches are optimized for 500 iterations to achieve target loss convergence (fewer iterations are used due to the absence of the curriculum required for the obtrusiveness term).  We then evaluate on the same set of held-out test images and record attack success rates. 

\begin{figure}[t!]
    \centering
    \begin{tabular}{c}
       \includegraphics[width=.98\columnwidth]{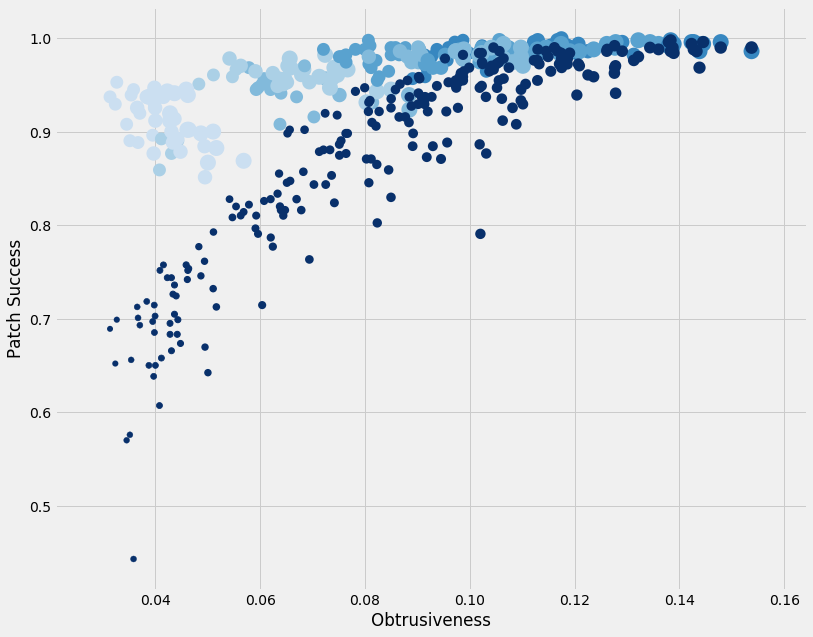}
    \end{tabular}
    \captionof{figure}{Partial patch results aggregated over the five top, middle, and bottom target labels each. Light colors imply greater transparency, larger dots imply larger patches. Results demonstrate that the larger, semi-transparent patches can be optimized which perform better than their fully opaque counterpart at the same obtrusiveness operating point.}
    \label{fig:all_labels_partial}
\end{figure}

\begin{figure*}[t!]
    \centering
    \begin{tabular}{ccc}
       \includegraphics[width=0.33\linewidth]{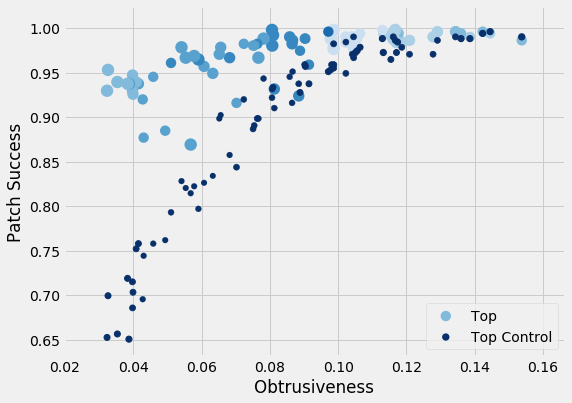}
       \includegraphics[width=0.33\linewidth]{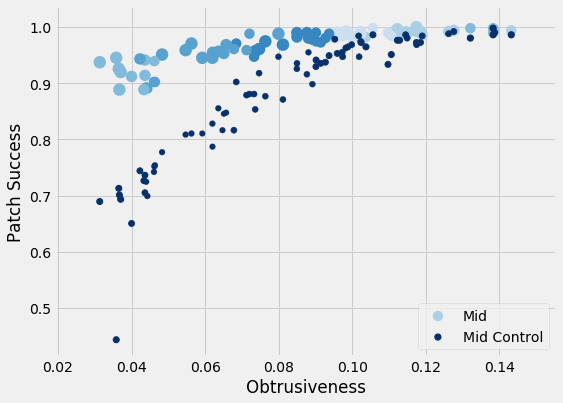}
       \includegraphics[width=0.33\linewidth]{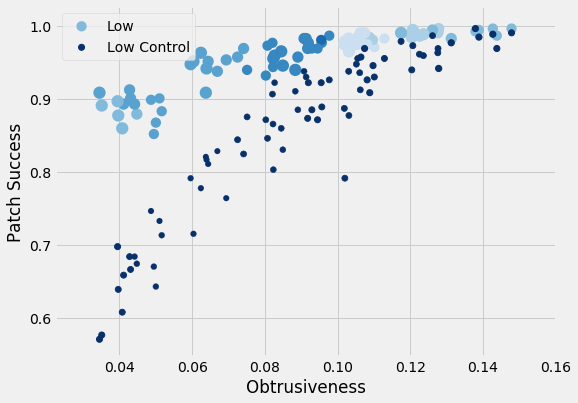}
    \end{tabular}
    \captionof{figure}{Partial patch results for the five top (left), middle (center), and bottom (right) performing target labels. Dot size is proportional to patch size; dot transparency is proportional to patch transparency. In all cases, the obtrusiveness-optimized patches demonstrate better performance overall compared to the fully opaque control patches.}
    \label{fig:partial_results_breakdown}
\end{figure*}

Results for these experiments are captured in Figures~\ref{fig:all_labels_partial} and~\ref{fig:partial_results_breakdown}.

\subsection{Discussion}
The results from Section~\ref{sub:obtrusive_exp} give us several insights into the trade-off between obtrusiveness and attack success. We demonstrate that we can generate patches that achieve similar or better success rates with lower obtrusiveness values.  
In particular, we see that for a range of operating points (i.e., obtrusiveness values), we can generate larger patches that achieve higher success but are more subtle relative to the standard patch attacks of the same size.  This is the first such demonstration of semi-transparent patches which are capable of achieving this kind of result.

However, we still recognize several potential avenues for improvement.  First, while we're the first to define a notion of patch attack obtrusiveness, we have not yet tied it to a true perceptual measure (necessary if we hope to claim to avoid detection by a human).  Second, we have constructed these experiments under rigid assumptions about rotation and location but could expand them in future iterations.  We believe our assumptions are reasonable given some of the likely physical attack use cases (e.g., patches on street signs).  Additionally, we would reiterate that we design these experiments for the general attack use case but could incorporate context and scene details in the optimization if we wish to address the scene-specific scenario.  Lastly, as in the experiments in Section~\ref{sec:EoT_experiments}, we have not fully explored nor explained the impact of target label on patch success.  We observe a clear impact and can hypothesize about why certain labels do better (e.g., due to scale/deformability/uniformity), but further investigation is necessary to better characterize these relationships.




\section{Conclusions}
This paper makes two key contributions: (1) we study a method and provide results that characterize the impact of real-world transformations on patch attack optimization and success, (2) we present a novel objective for the design of ideal patches that trade obtrusiveness and success rate, and develop companion approaches for optimization of patch obtrusiveness for producing more subtle patch attacks without sacrificing performance.  We have demonstrated that the results of (1) can be directly leveraged to improve patch attacks (e.g., as shown with (2)).  We believe that the methodology and results presented in this paper will help provide greater guidance and focus in the community in understanding patch attack capabilities and limitations.  Lastly, our obtrusiveness measure led to the production of semi-transparent attacks (for the first time) and opens the door for exploring many other variants of this approach.
With these results we want to underscore the importance of generating these attacks (and subsequent defenses) not as a means for defeating visual recognition systems, but rather as a way to improve understanding of the robustness of these systems and gain greater insight into their inner workings and possible defenses, which is left for future endeavors.



{\small
\bibliographystyle{ieee_fullname}
\bibliography{egbib}
}

\end{document}